\documentclass{article}

\usepackage[preprint]{neurips_2025}

\usepackage[utf8]{inputenc} 
\usepackage[T1]{fontenc}    
\usepackage{hyperref}       
\usepackage{url}            
\usepackage{booktabs}       
\usepackage{amsfonts}       
\usepackage{nicefrac}       
\usepackage{microtype}      
\usepackage{xcolor}         
\usepackage{graphicx}
\usepackage{wrapfig}
\usepackage{listings}
\usepackage{multirow}
\usepackage{multicol}
\usepackage{pifont} 
\usepackage[most]{tcolorbox}

\tcbuselibrary{listingsutf8}
\newtcblisting{pythoncodebox}{
  colback=gray!5!white,    
  colframe=gray!75!black,  
  listing only,
  listing options={
    language=Python,
    basicstyle=\ttfamily\small,
    keywordstyle=\color{blue},
    commentstyle=\color{gray},
    stringstyle=\color{red},
    showstringspaces=false,
    breaklines=true
  },
  left=5pt,
  right=5pt,
  top=5pt,
  bottom=5pt,
  enhanced,
  boxrule=0.5pt,
  arc=2mm,
  drop shadow
}
\newtcolorbox{infobox}[2][]{ 
  enhanced,
  breakable,
  colback=gray!5!white,
  colframe=gray!75!black,
  boxrule=0.5pt,
  arc=2mm,
  left=5pt,
  right=5pt,
  top=5pt,
  bottom=5pt,
  drop shadow,
  title={#2},
  fonttitle=\bfseries,
  #1
}
\title{When LLM Meets Time Series: Can LLMs Perform Multi-Step Time Series Reasoning and Inference}

\author{%
  Wen Ye, Jinbo Liu, Defu Cao, Wei Yang, Yan Liu\\
  University of Southern California\\
  \texttt{\{yewen,jinboliu,defucao,wyang930,yanliu.cs\}@usc.edu} \\
}
\newcommand{\modelname}{TSAIA}
\newcommand{\modelnamespace}{TSAIA }

\newcommand{\odc}{\textsuperscript{\dag}} 
\newcommand{\tos}{\textsuperscript{\ddag}}
\begin{document}

\maketitle

\begin{abstract}

The rapid advancement of Large Language Models (LLMs) has sparked growing interest in their application to time series analysis tasks. However, their ability to perform complex reasoning over temporal data application domain remains significantly underexplored. To achieve the goal, one first step is to establish a rigorous benchmark dataset for evaluation. In this work, we introduce \modelnamespace Benchmark, a first attempt to evaluate LLMs as time series artificial intelligence assistant. To ensure both scientific rigor and practical relevance, we surveyed over 20 academic publications and identified 33 real-world task formulations. The benchmark encompasses a broad spectrum of challenges, ranging from constraint-aware forecasting to anomaly detection with threshold calibration—tasks that require compositional reasoning and multi-step time series analysis. The question generator is designed to be dynamic and extensible, supporting continuous expansion as new datasets or task types are introduced. Given the heterogeneous nature of the tasks, we adopt task-specific success criteria and tailored inference quality metrics to ensure meaningful evaluation for each task. We apply this benchmark to assess 8 state-of-the-art LLMs under a unified evaluation protocol. Our analysis reveals limitations in current models' ability to assemble complex time series analysis workflows, underscoring the need for specialized methodologies for adaptation towards domain-specific applications. Our benchmark is available at \url{https://huggingface.co/datasets/Melady/TSAIA}, and the code is available at \url{https://github.com/USC-Melady/TSAIA}. 
\end{abstract}

\section{Introduction}
Large Language Models (LLMs) have demonstrated remarkable general-purpose capabilities across a range of domains, including language understanding \cite{dong2019unified}, code generation \cite{jiang2024survey}, and scientific reasoning \cite{taylor2022galactica}. Despite their growing capabilities, large language models have not been evaluated for time series analysis, which constitutes a fundamental competency for data analysts and scientists across critical domains including energy \cite{alvarez2010energy}, finance \cite{sezer2020financial}, climate science \cite{mudelsee2019trend}, and healthcare \cite{rathlev2007time}. In practice, time series workflows in the real world are inherently complex \cite{yan2021learning, han2021real}: they demand multi-step reasoning \cite{fu2022complexity}, precise numerical computation \cite{cvejoski2022future}, integration of domain knowledge \cite{xue2024domain}, and adherence to operational constraints \cite{wang220kv}. With the advent of agent systems, there is growing interest among practitioners in developing intelligent agents capable of interpreting natural language instructions to perform time series analysis. However, given that time series understanding poses challenges even for humans, researchers are striving to establish a clear paradigm for how LLMs can effectively contribute to these complex downstream tasks. A crucial step toward empowering LLMs in time series applications is the development of rigorous benchmark datasets for comprehensive evaluation.

Despite recent interest in evaluating LLMs' ability to solve temporal-related tasks \cite{merrill2024language}, existing benchmarks are insufficient for evaluating whether LLMs can function as general-purpose time series inference agents capable of constructing complete, constraint-aware analytical workflows. Most focus on individual tasks \cite{du2024tsi} and often evaluated under fixed experiment configuration (using 336 historical timesteps to predict the next 96 timesteps \cite{zhou2021informer}), direct QA questions that only involve temporal concepts but do not involve time series or require numerical precision ability \cite{wang2024tram}, and lack of coverage on real-world operational constraints. Critically, they fail to capture whether LLMs can serve as a general-purpose time series assistant.   

To push the boundary of LLM-based time series agents and address the lack of a suitable benchmark, we introduce \textbf{T}ime \textbf{S}eries \textbf{A}rtificial \textbf{I}ntelligence \textbf{A}ssistant \textbf{\modelnamespace} Benchmark. The benchmark is grounded in practical relevance (\textit{what tasks are suitable?}), enables dynamic extensibility (\textit{how to generate specific task instances?}), and facilitates unified evaluation (\textit{how to perform evaluation over heterogeneous task types?}). We collected and curated diverse datasets from energy, climate science, finance, and healthcare domains. Drawing from a review of over 20 time series application publications, identified 33 common tasks types adding up to 1054 questions in total, covering predictive, diagnostic, analytical, and decision-making tasks. To achieve good performance on \modelname, the following capabilities are needed: compositional reasoning \cite{li2024understanding} (the sequential execution of logical and numerical operations to construct end-to-end analytical pipelines), comparative reasoning (selecting the optimal asset based on calculated summary indicators), commonsense reasoning \cite{davis2015commonsense} (identifying plausible covariates for the target variable), decision-oriented reasoning (interpreting risk-return metrics in investment contexts), and numerical precision. 

We evaluate eight state-of-the-art models: GPT-4o \cite{hurst2024gpt}, Qwen2.5-Max \cite{qwen2025max}, Llama-3.1 Instruct 70B \cite{grattafiori2024llama}, Claude-3.5 Sonnet \cite{anthropic2024claude35sonnet}, DeepSeek \cite{liu2024deepseek}, Gemini-2.0 \cite{google2024gemini2}, Codestral \cite{mistral2024codestral}, and DeepSeek-R \cite{guo2025deepseek}—using the CodeAct agent framework \cite{wang2024executable} implemented via AgentScope \cite{agentscope}. Each agent is prompted to generate executable Python code to perform task-specific inference, with iterative refinement based on execution feedback. This approach accommodates LLM limitations on premature output termination for long sequence output \cite{song2025hansel} and tokenizing numerical values to discrete tokens \cite{spathis2024first}. Our findings reveal that while certain models show strengths on narrow task types, none reliably generalizes across the full benchmark. Common failure modes include inadequate numerical result/trivial prediction and the inability to assemble complex workflows. These results underscore the challenges of structured numerical reasoning in real-world time series settings and position \modelnamespace as a critical benchmark for future development of domain-grounded, reasoning-capable time series AI assistants.

\begin{figure}
    \centering
    \includegraphics[width=\linewidth]{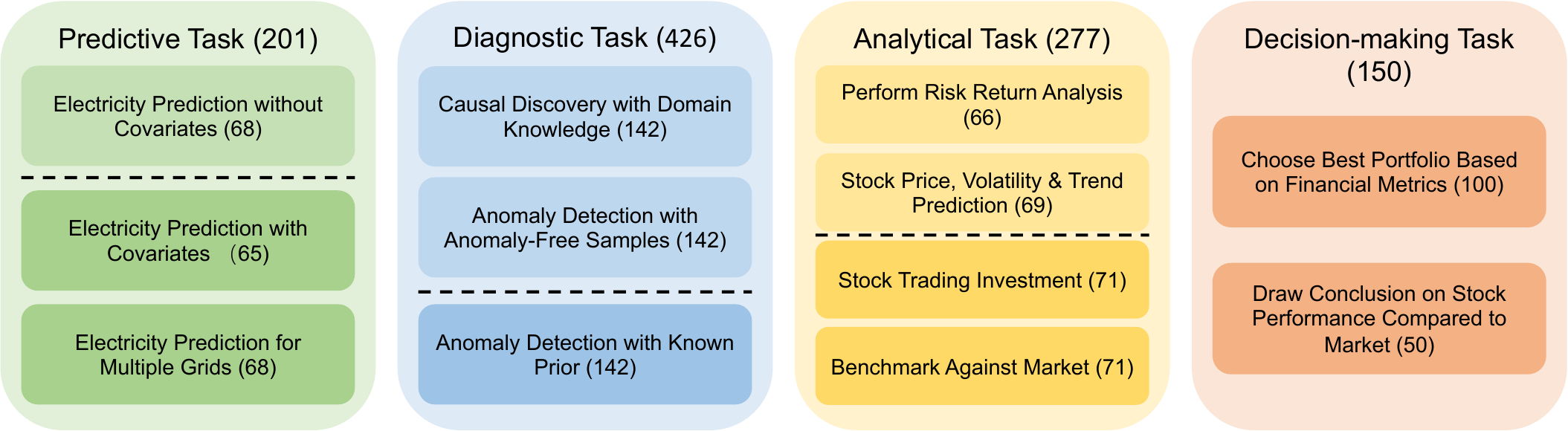}
    \caption{Categorization of Tasks in \modelname. Lighter colors denote tasks with less difficulty and darker colors denote tasks with higher difficulty.}
    \label{fig:task annotation}
\end{figure}
\section{\modelnamespace Benchmark}

\subsection{Task Annotation\& Data Collection: What time series tasks hold real world practicality?}

To evaluate time series AI assistants effectively, we focus on tasks grounded in real-world use cases that data analyst in different time series application domains may face. We aim for the tasks to assess an assistant's ability to handle practical scenarios involving time series data, ensuring their utility in everyday analytical workflows. We start by forming a group of three graduate students with backgrounds in machine learning and applied time series analysis to collect real world applications of time series analysis that exhibit multi-step complexity and requires reasoning. We reviewed over 20 research publications addressing time series analysis challenges, extracting and converting core problems into benchmark tasks. The collected tasks can be categorized into four broad groups based on the nature of the underlying problem as shown in figure \ref{fig:task annotation}: \textbf{(1) Predictive Tasks} involve forecasting with or without covariates, with constraint-aware prediction that must comply with real-world operational requirements such as maximum/minimum load limits \cite{wang220kv,greif1999short}, ramp rate constraints \cite{schaible2024application,gkavanoudis2023provision}, and variability thresholds \cite{han2021real}. \textbf{(2) Diagnostic Tasks} focus on identifying abnormal patterns or latent structures within the data. This includes anomaly detection tasks, which may leverage reference samples or known anomaly frequencies \cite{yan2021learning,song2024unsupervised,niu2020lstm,verma2023deep}; and causal discovery tasks, where models infer binary causal graphs from observational time series with domain-specific priors such as partial causal ratios or qualitative relationships \cite{aoki2020parkca}. \textbf{(3) Analytical Tasks} target analytics based on time series trends, particularly within financial domains. These tasks may require performing risk-return analysis \cite{riley2022maximum,miglietti2020bitcoin} or generating trading strategies using historical asset prices \cite{shalini2019picking,pramudya2020technical}. \textbf{(4) Decision-Making Tasks} are framed as multiple-choice questions, primarily in the financial domain. These require models to analyze structured summaries such as portfolio performance under different metrics \cite{steinki2015common} or compare a stock’s performance to a market index \cite{gupta1999information}. The correct answer involves selecting the option that best aligns with the target financial principle, testing both computation and reasoning \cite{legrenzi1993focussing}.
 
 For each task, we manually craft task templates based on extracted problem formulations, collect and preprocess the relevant datasets, and, where appropriate, appended domain-specific metadata. Time series datasets in \modelnamespace are collected from publicly available repositories spanning domains including energy management, climate science, finance, and healthcare. Specifically, it incorporates electricity grid load data, solar and wind power measurements with corresponding weather covariates \cite{zheng2021psml}\footnotemark, building energy usage patterns \cite{LEAD2022}\footnotemark[1], ERA5 climate variables \cite{hersbach2020era5}\footnotemark[1], MIT-BIH ECG signals \cite{goldberger2000physiobank}\odc, and economic indicators such as stock indices and individual stock prices \cite{yahoo_finance}\tos.

\footnotetext{ CC BY 4.0 licence; 
\odc\ Open Data Commons Attribution 1.0 licence.\\
\tos\ Yahoo Finance Terms of Service: personal, non-commercial use only.}

\begin{figure*}
    \centering
    \includegraphics[width=\linewidth]{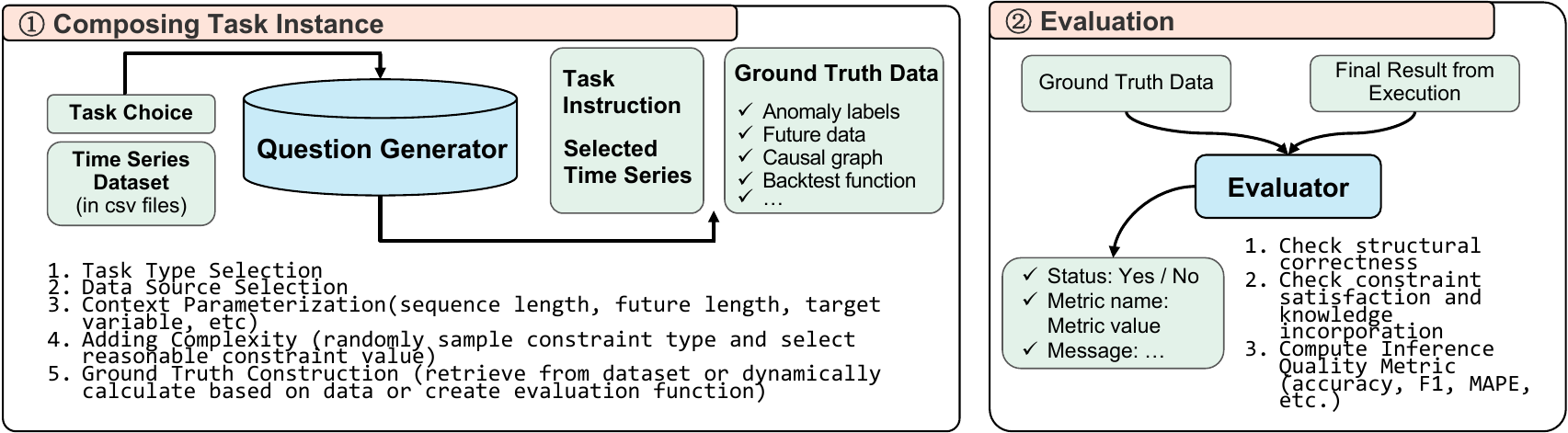}
    \caption{The proposed pipeline for multi-step time series inference task instance generation and evaluation protocol. }
    \label{fig:evaluation}
    \vspace{-0.15in}
\end{figure*}
\begin{figure}
    \centering
    \includegraphics[width=\linewidth]{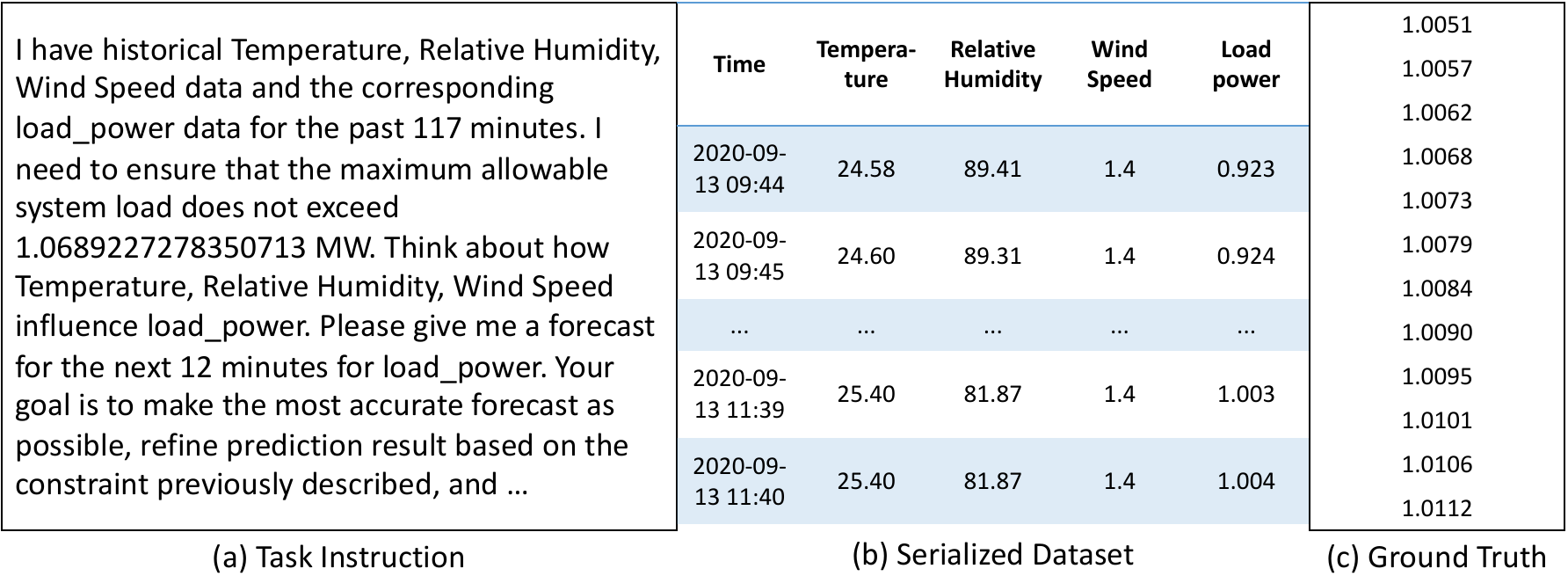}
    \caption{Example Task Instance containing the task instruction, accompanied serialized dataset, and ground truth.}
    \label{fig:example} 
\end{figure}
\subsection{Question Generator: How to generate specific task instances?}

We design a modular and programmatic pipeline (illustrated in figure \ref{fig:evaluation}) to generate diverse task instances for evaluating time series inference tasks. The generation process proceeds in five steps: \textit{(1) Task Type Selection}: The generation process begins by selecting a task type from a predefined library. \textit{(2) Data Source Selection}: A time series dataset is sampled from a repository of CSV datasets. \textit{(3) Context Parameterization}: Once a dataset is selected, the generator randomly configures instance-specific parameters such as input sequence length, forecasting horizon (if applicable), target variable, covariate inclusion. These contextual parameters are crucial for defining the scope of the task instance. The natural language template corresponding to the selected task type is dynamically populated with the sampled parameters and contextual information. \textit{(4) Adding Complexity}: To reflect real-world conditions, the generator samples domain-specific constraints or auxiliary knowledge defined by the task template and adds to the task instruction. \textit{(5) Ground Truth Construction}: Finally, ground truth solutions are either directly retrieved from the dataset, dynamically computed based on the task parameters, or delegated to domain-specific evaluation functions. This step ensures that each instance is paired with an executable reference output, enabling robust automatic evaluation.

As shown in figure \ref{fig:example}, each task instance contains a natural language instruction paired with structured time series inputs and corresponding ground truth data. By design, the benchmark framework is extensible and dynamic. New task instances can be generated automatically by applying the same pipeline to additional time series data sources when accompanied by its designated task template, supporting ongoing evaluation and adaptation to new domains. This supports long-term benchmarking efforts and enables ongoing expansion across domains.

\subsection{Evaluation: How to perform evaluation on heterogeneous task instances?}

\begin{table}[ht]
\centering

\resizebox{\textwidth}{!}{
\begin{tabular}{ p{0.25\textwidth}   p{0.6\textwidth}  p{0.2\textwidth} }
\hline
\textbf{Task Type} & \textbf{Success Criterion} & \textbf{Metrics} \\
\hline
Constrained Forecasting & Prediction is of correct shape and  satisfies the specified operational constraint and the prediction is non-trivial (MAPE<1) & MAPE \\

\hline
Anomaly Detection w/ reference samples & A binary sequence with correct length is obtained and the prediction is non-trivial (F1-score>0) & F1-score \\
\hline
Causal Inference w/ domain knowledge & A binary causal matrix with correct shape is returned. The provided domain knowledge is incorporated & Accuracy\\
\hline
Financial Analytics & A scalar value is returned and the prediction is non-trivial (absolute error<0.05) & Absolute Error\\
\hline
Financial Trading & An investment signal of correct length is returned and there is no loss in investment &  CR, AR, MDD\\
\bottomrule
\end{tabular}
}
\caption{Task-specific success criteria and inference quality metrics. CR denotes Cumulative Return, AR denotes Annualized Return, MDD denotes Maximum Drawdown.}
\label{tab:success_criteria}
\end{table}

\paragraph{Ground Truth Data}
Ground truth labels are defined based on the task type and the nature of the expected output. For questions demanding numerical outputs, the ground truth is obtained through one of three mechanisms. \textit{(1) Retrieved from dataset}:
For predictive tasks, the ground truth corresponds to the future values of the target variable over the specified forecasting horizon. In diagnostic tasks such as anomaly detection, binary anomaly labels serve as ground truth for the given time window. \textit{(2) Dynamically computed based on data}: In risk \& return analysis, ground truth values are derived based on predefined formulas for key performance metrics and are dynamically computed based on data. \textit{(3) Defined by an evaluation function}: In contrast, trading tasks do not have a unique ground truth strategy, as multiple plausible investment approaches may yield positive returns. Instead of comparing predicted strategies to a reference solution, we evaluate the generated investment signals through backtesting on held-out stock price data over the hypothetical investment period. Performance is assessed using financial metrics such as cumulative return \cite{clarke2011minimum}, annualized return \cite{booth1992diversification}, and maximum drawdown \cite{magdon2004maximum}.

For multiple-choice questions, the answer is determined through quantitative evaluation. In portfolio-based questions, each option corresponds to a distinct portfolio constructed from sampled stocks and associated holding ratios, and the correct answer is the portfolio that optimizes the target metric such as Sharpe ratio \cite{sharpe1994sharpe}, Value at Risk \cite{duffie1997overview} based on selected data. In stock-versus-market comparisons, ground truth relationship between the stock and market is derived from computed indicators Jensen's alpha and beta \cite{fisher2018risk}. To avoid positional bias, correct answers are randomly assigned to lettered choices, ensuring a uniform distribution across answer positions throughout the dataset.

\begin{table}[!htbp]
\centering

\begin{tabular}{lccccc}
\toprule
\textbf{Benchmark} & \textbf{Dynamic} & \textbf{TS involved} & \textbf{Reasoning} & \textbf{\#Tasks} & \textbf{Task Type} \\\hline
Test of Time~\cite{fatemi2024test}            & \ding{55}  & \ding{55} & \ding{51} & 1 & QA \\
TRAM~\cite{wang2024tram}                      & \ding{55}  & \ding{55} & \ding{51} & 1 & QA \\
\hline
TSI-Bench~\cite{du2024tsi}                    & \ding{55}  & \ding{51} & \ding{55} & 1 & TS Analysis \\
TSB-AD~\cite{liu2024the}                      & \ding{55}  & \ding{51} & \ding{55} & 1 & TS Analysis \\
GIFT-Eval~\cite{aksugift}                     & \ding{55}  & \ding{51} & \ding{55} & 1 & TS Analysis \\
TFB~\cite{qiu2024tfb}                         & \ding{55}  & \ding{51} & \ding{55} & 1 & TS Analysis \\
Time-MMD~\cite{liu2024timemmd}                & \ding{55}  & \ding{51} & \ding{55} & 1 & TS Analysis \\
CiK~\cite{williams2024context}                & \ding{55}  & \ding{51} & \ding{55} & 1 & TS Analysis \\
TGTSF~\cite{xu2024beyond}                     & \ding{55}  & \ding{51} & \ding{55} & 1 & TS Analysis \\
\hline
LLM TS Struggle~\cite{merrill2024language}     & \ding{55}  & \ding{51} & \ding{51} & 2 & QA, TS Analysis \\

MTBench~\cite{chen2025mtbench}                & \ding{55}  & \ding{51} & \ding{51} & 3 & QA, TS Analysis \\
ChatTime~\cite{chatime}                       & \ding{55}  & \ding{51} & \ding{51} & 3 & QA, TS Analysis \\ \hline
\textbf{\modelname (Ours) }                                         & \ding{51}  & \ding{51} & \ding{51} & \textbf{4} & QA, TS Analysis \\ \bottomrule
\end{tabular}%
\caption{Comparison of \modelnamespace and existing temporal-related benchmarks. Dynamic indicates whether new task instances can be continuously generated.}
\label{tab:ts-benchmarks-final}
\end{table}

\paragraph{Evaluation Criteria}
Given the diverse nature of the tasks—ranging from numerical forecasting to causal inference and constraint satisfaction—it is neither appropriate nor effective to rely on a single evaluation metric. Instead, evaluation must be tailored to each task to ensure a meaningful assessment of model performance. Across all tasks, outputs must conform to expected formats, satisfy any injected constraints, and appropriately incorporate provided domain knowledge. Trivial or degenerate outputs such as poor quality predictions or all-zero anomaly labels are flagged as failures, even if their format is technically correct. Each task type is governed by well-defined success criteria and inference quality metrics, designed to reflect practical expectations of correctness and utility. Table~\ref{tab:success_criteria} summarizes the criteria used for each task.

The evaluation process follows a three-stage protocol. In the first stage, outputs are validated for structural correctness, shape conformity. The second stage checks against specified constraint and domain knowledge incorporation. Lastly, the inference quality metric is computed relative to ground truth data. Results are returned in a structured format, including the success status, diagnostic messages, and detailed metric scores. Failures are categorized into execution errors (model output fails to run or parse), constraint violations (outputs violate injected domain-specific rules), and low quality outputs (predictions meet format expectations but fall short on metric thresholds). Multiple choice questions have a simple evaluation procedure of checking against ground truth letter option.

\subsection{Comparison with Existing Benchmarks}
As shown in table ~\ref{tab:ts-benchmarks-final}, existing time-series benchmarks fall into three main groups, each lacking one or more component for evaluating time series AI assistant:
First, datasets such as Test of Time~\cite{fatemi2024test} and TRAM~\cite{wang2024tram} present pure QA tasks (ordering, duration, arithmetic) with no time-series included (TS involved:\ding{55}). While they evaluate logical reasoning, they cannot test how models process numerical signals. Secondly, 
benchmarks like TSI-Bench~\cite{du2024tsi}, TSB-AD~\cite{liu2024the}, GIFT-Eval~\cite{aksugift}, TFB~\cite{qiu2024tfb}, Time-MMD~\cite{liu2024timemmd}, CiK~\cite{williams2024context}, and TGTSF~\cite{xu2024beyond} focus on a single static time‐series analysis including imputation, anomaly detection, and forecasting over fixed datasets with pre-defined sliding window size for evaluation (Dynamic:\ding{55}, Reasoning:\ding{55}, Tasks=1). Lastly, recent efforts on hybrid QA and analysis such as  MTBench~\cite{chen2025mtbench}, ChatTime~\cite{chatime} combine time-series and text inputs (TS involved:\ding{51}) and include reasoning components (Reasoning:\ding{51}), yet remain fixed setting (Dynamic:\ding{55}) and only covers context-aided forecasting in time series analysis component part. \modelnamespace arises as first of its kind time series inference benchmark with practical relevance, task diversity, and supports continuous expansion.

\section{Experiments}

\begin{table}[htbp]
\centering

\resizebox{\textwidth}{!}{%
\begin{tabular}{llcccccccc}
\toprule
\multicolumn{2}{c}{\textbf{Metric}} & \textbf{GPT-4o} & \textbf{Qwen-Max} & \textbf{Llama3.1} & \textbf{Claude-3.5} & \textbf{DeepSeek} & \textbf{Gemini-2.0} & \textbf{Codestral} & \textbf{DeepSeek-R} \\
\midrule
\multicolumn{10}{l}{\textbf{Electricity Prediction with Covariates}} \\
\midrule
\multirow{2}{*}{Max Load} & Success Rate & 0.50 & 0.75 & 0.56 & 0.88 & \textcolor{red}{1.00} & 0.19 & 0.31 & \textcolor{blue}{0.94} \\
 & MAPE (std) & 0.09 (0.12) & 0.07 (0.06) & 0.10 (0.11) & 0.11 (0.12) & \textcolor{red}{0.10 (0.12)} & 0.52 (0.41) & 0.03 (0.03) & \textcolor{blue}{0.10 (0.11)} \\
\midrule
\multirow{2}{*}{Min Load} & Success Rate & 0.76 & 0.82 & 0.65 & 0.82 & \textcolor{blue}{0.88} & 0.18 & 0.59 & \textcolor{red}{1.00} \\
 & MAPE (std) & 0.11 (0.11) & 0.09 (0.11) & 0.10 (0.11) & 0.09 (0.11) & \textcolor{blue}{0.12 (0.18)} & 0.09 (0.04) & 0.09 (0.10) & \textcolor{red}{0.09 (0.11)} \\
\midrule
\multirow{2}{*}{Load Ramp Rate} & Success Rate & 0.46 & 0.80 & 0.53 & \textcolor{blue}{0.93} & 0.80 & 0.13 & 0.47 & \textcolor{red}{0.93} \\
 & MAPE (std) & 0.18 (0.14) & 0.14 (0.12) & 0.15 (0.07) & \textcolor{blue}{0.19 (0.19)} & 0.11 (0.08) & 0.04 (0.01) & 0.12 (0.08) & \textcolor{red}{0.11 (0.07)} \\
\midrule
\multirow{2}{*}{Load Variability} & Success Rate & 0.47 & \textcolor{blue}{0.76} & 0.29 & 0.29 & 0.76 & 0.06 & 0.35 & \textcolor{red}{0.94} \\
 & MAPE (std) & 0.20 (0.31) & \textcolor{blue}{0.13 (0.16)} & 0.09 (0.12) & 0.09 (0.12) & 0.19 (0.27) & 0.05 (0.00) & 0.04 (0.03) & \textcolor{red}{0.11 (0.14)} \\
\midrule
\multicolumn{10}{l}{\textbf{Electricity Prediction without Covariates}} \\
\midrule
\multirow{2}{*}{Max Load} & Success Rate & 1.00 & 0.94 & 0.94 & \textcolor{blue}{1.00} & 0.94 & 0.41 & \textcolor{red}{1.00} & 0.71 \\
 & MAPE (std) & 0.18 (0.16) & 0.10 (0.07) & 0.16 (0.10) & \textcolor{blue}{0.15 (0.13)} & 0.15 (0.12) & 0.10 (0.02) & \textcolor{red}{0.12 (0.07)} & 0.23 (0.26) \\
\midrule
\multirow{2}{*}{Min Load} & Success Rate & \textcolor{blue}{0.94} & \textcolor{red}{0.94} & 0.94 & 0.94 & 0.88 & 0.29 & 0.71 & 0.88 \\
 & MAPE (std) & \textcolor{blue}{0.14 (0.08)} & \textcolor{red}{0.14 (0.08)} & 0.17 (0.08) & 0.17 (0.09) & 0.13 (0.09) & 0.12 (0.03) & 0.14 (0.05) & 0.17 (0.16) \\
\midrule
\multirow{2}{*}{Load Ramp Rate} & Success Rate & 0.76 & \textcolor{red}{1.00} & 0.71 & 0.76 & 0.82 & 0.24 & \textcolor{blue}{0.88} & 0.76\\
 & MAPE (std) & 0.24 (0.19) & \textcolor{red}{0.23 (0.22)} & 0.21 (0.11) & 0.28 (0.16) & 0.19 (0.20) & 0.42 (0.28) & \textcolor{blue}{0.29 (0.30)} & 0.22 (0.13) \\
\midrule
\multirow{2}{*}{Load Variability} & Success Rate & 0.82 & \textcolor{red}{0.88} & \textcolor{blue}{0.82} & 0.65 & 0.76 & 0.41 & 0.71 & 0.65 \\
 & MAPE (std) & 0.17 (0.12) & \textcolor{red}{0.13 (0.09)} & \textcolor{blue}{0.15 (0.09)} & 0.16 (0.12) & 0.19 (0.17) & 0.39 (0.39) & 0.13 (0.07) & 0.24 (0.24) \\
\midrule
\multicolumn{10}{l}{\textbf{Electricity Prediction for Multiple Grids}} \\
\midrule
\multirow{2}{*}{Max Load} & Success Rate & 0.76 & {0.88} & 0.47 & \textcolor{blue}{0.88} & \textcolor{red}{0.94} & 0.47 & 0.12 & 0.88 \\
 & MAPE (std) & 0.21 (0.27) & {0.21 (0.24)} & 0.64 (0.31) & \textcolor{blue}{0.18 (0.20)} & \textcolor{red}{0.16 (0.21)} & 0.34 (0.39) & 0.10 (0.03) & 0.23 (0.27) \\
\midrule
\multirow{2}{*}{Min Load} & Success Rate & 0.76 & 0.88 & 0.24 & \textcolor{blue}{0.94} & \textcolor{red}{0.94} & 0.18 & 0.29 & 0.94 \\
 & MAPE (std) & 0.10 (0.12) & 0.18 (0.29) & 0.46 (0.37) & \textcolor{blue}{0.13 (0.20)} & \textcolor{red}{0.08 (0.11)} & 0.01 (0.01) & 0.23 (0.37) & 0.16 (0.23) \\
\midrule
\multirow{2}{*}{Load Ramp Rate} & Success Rate & 0.65 & 0.65 & 0.88 & 0.88 & \textcolor{blue}{0.94} & 0.29 & 0.29 & \textcolor{red}{1.00} \\
 & MAPE (std) & 0.19 (0.24) & 0.18 (0.18) & 0.73 (0.33) & 0.27 (0.21) & \textcolor{blue}{0.21 (0.21)} & 1.00 (0.00) & 0.10 (0.05) & \textcolor{red}{0.19 (0.19)} \\
\midrule
\multirow{2}{*}{Load Variability} & Success Rate & 0.41 & \textcolor{blue}{0.59} & 0.53 & 0.41 & \textcolor{red}{0.59} & 0.35 & 0.29 & 0.53 \\
 & MAPE (std) & 0.15 (0.13) & \textcolor{blue}{0.18 (0.23)} & 0.61 (0.36) & 0.11 (0.13) & \textcolor{red}{0.18 (0.14)} & 0.90 (0.23) & 0.19 (0.13) & 0.24 (0.25) \\
\bottomrule
\end{tabular}%
}
\caption{Model Performance on Predictive Task. \textcolor{red}{Red} indicates best result, \textcolor{blue}{Blue} indicates second best.}
\label{tab:predictive}
\end{table}

\begin{table}[htbp]
\centering

\resizebox{\textwidth}{!}{%
\begin{tabular}{llcccccccc}
\toprule
\multicolumn{2}{c}{\textbf{Benchmark}} & \textbf{GPT-4o} & \textbf{Qwen-Max} & \textbf{Llama3.1} & \textbf{Claude-3.5} & \textbf{DeepSeek} & \textbf{Gemini-2.0} & \textbf{Codestral} & \textbf{DeepSeek-R} \\
\midrule
\multicolumn{10}{l}{\textbf{Diagnostic Task w/ Reference Samples}} \\
\midrule
{Extreme Weather Detection} & Success Rate & 0.24 & 0.23 & 0.23 & \textcolor{red}{0.62} & 0.23 & 0.14 & 0.23 & \textcolor{blue}{0.34} \\
 w/ Reference Samples & F1 (std) & 0.91 (0.23) & 0.90 (0.24) & 0.90 (0.24) & \textcolor{red}{0.90 (0.18)} & 0.90 (0.24) & 0.96 (0.06) & 0.91 (0.23) & \textcolor{blue}{0.91 (0.20)} \\
\midrule

{ECG Signal Anomaly} & Success Rate & 0.51 & 0.17 & 0.55 & \textcolor{red}{0.68} & 0.54 & 0.10 & 0.59 & \textcolor{blue}{0.63} \\
 w/ Reference Samples& F1 (std) & 0.55 (0.35) & 0.70 (0.29) & 0.43 (0.36) & \textcolor{red}{0.65 (0.32)} & 0.54 (0.34) & 0.01 (0.00) & 0.58 (0.34) & \textcolor{blue}{0.61 (0.32)} \\

\midrule
\multicolumn{10}{l}{\textbf{Causal Discovery w/ Domain Knowledge}} \\
\midrule
{Causal Discovery w/} & Success Rate & 0.94 & 0.92 & \textcolor{blue}{0.99} & \textcolor{red}{1.00} & 0.97 & 0.39 & 0.94 & {0.96} \\
Quantitative Knowledge& Accuracy (std) & 0.69 (0.09) & 0.77 (0.11) & \textcolor{blue}{0.78 (0.12)} & \textcolor{red}{0.77 (0.09)} & 0.71 (0.11) & 0.42 (0.18) & 0.72 (0.11) & {0.77 (0.10)} \\
\midrule
{Causal Discovery w/} & Success Rate & 0.85 & 0.70 & 0.83 & \textcolor{blue}{0.97} & 0.96 & 0.45 & 0.93 & \textcolor{red}{0.99} \\
Qualitative Knowledge & Accuracy (std) & 0.87 (0.17) & 0.79 (0.17) & 0.77 (0.18) & \textcolor{blue}{0.89 (0.16)} & 0.89 (0.14) & 0.72 (0.20) & 0.88 (0.15) & \textcolor{red}{0.90 (0.12)} \\
\midrule
\multicolumn{10}{l}{\textbf{Anomaly Detection across Multiple Sequences}} \\

\midrule
{Extreme Weather Detection} & Success Rate & 0.87 & 0.31 & 0.03 &  0.97& \textcolor{blue}{0.97}  & 0.37 & 0.23 & \textcolor{red}{1.00} \\
 w/ Known Anomaly Rate& F1 (std) & 0.53 (0.25) & 0.62 (0.19) & 0.68 (0.05) & {0.73 (0.12)} & \textcolor{blue}{0.72 (0.11)} & 0.65 (0.18) & 0.42 (0.31) & \textcolor{red}{0.72 (0.10)} \\
\midrule
{Energy Usage Anomaly} & Success Rate & 0.87 & 0.52 & 0.77 & \textcolor{blue}{1.00} & \textcolor{red}{1.00} & 0.23 & 0.58 & 0.96 \\
 w/ Known Anomaly Rate& F1 (std) & 0.08 (0.09) & 0.14 (0.20) & 0.15 (0.11) & \textcolor{blue}{0.48 (0.18)} & \textcolor{red}{0.50 (0.19)} & 0.19 (0.21) & 0.06 (0.06) & 0.40 (0.23) \\
\bottomrule
\end{tabular}%
}
\caption{Model Performance on Diagnostic Task. \textcolor{red}{Red} indicates best result, \textcolor{blue}{Blue} indicates second best.}
\label{tab:diagnostic}
\end{table}

\begin{table}[htbp]
\centering

\resizebox{\textwidth}{!}{%
\begin{tabular}{llcccccccc}
\toprule
\multicolumn{2}{c}{\textbf{Benchmark}} & \textbf{GPT-4o} & \textbf{Qwen-Max} & \textbf{Llama3.1} & \textbf{Claude-3.5} & \textbf{DeepSeek} & \textbf{Gemini-2.0} & \textbf{Codestral} & \textbf{DeepSeek-R} \\
\midrule
\multicolumn{10}{l}{\textbf{Stock Prediction}} \\
\midrule
\multirow{2}{*}{Future Price} & Success Rate & 0.96 & \textcolor{red}{1.00} & 0.70 & 0.74 & 0.87 & 0.17 & 0.39 & \textcolor{blue}{1.00} \\
 & MAPE (std) & 0.06 (0.08) & \textcolor{red}{0.05 (0.07)} & 0.06 (0.08) & 0.12 (0.16) & 0.05 (0.07) & 0.28 (0.42) & 0.05 (0.05) & \textcolor{blue}{0.05 (0.07)} \\
\midrule
\multirow{2}{*}{Future Volatility} & Success Rate & \textcolor{blue}{0.83} & 0.43 & 0.39 & \textcolor{red}{0.74} & 0.57 & 0.17 & 0.57 & 0.61 \\ 
& MAPE (std) & \textcolor{blue}{0.70 (0.28)} & 0.83 (0.26) & 0.64 (0.29) & \textcolor{red}{0.75 (0.31)} & 0.90 (0.13) & 0.84 (0.24) & 0.61 (0.32) & 0.77 (0.24) \\ 
\midrule 
\multirow{2}{*}{Future Trend} & Success Rate & {0.43} & 0.30 & \textcolor{red}{0.57} & {0.26} & 0.43 & 0.04 & \textcolor{blue}{0.52} & 0.35 \\ 
& Accuracy (std) & 0.90 (0.20) & 0.86 (0.23) & \textcolor{red}{0.88 (0.21)} & {1.00 (0.00)} & 0.85 (0.23) & 1.00 (0.00) & \textcolor{blue}{0.96 (0.14)} & 0.81 (0.24) \\
\midrule
\multicolumn{10}{l}{\textbf{Risk/Return Estimation}} \\
\midrule
\multirow{2}{*}{Annualized Return} & Success Rate & \textcolor{blue}{0.45} & -- & 0.09 & 0.27 & 0.36 & -- & 0.18 & \textcolor{red}{0.55} \\
 & Abs Error (std) & \textcolor{blue}{0.02 (0.02)} & -- & 0.01 (0.00) & 0.01 (0.01) & 0.02 (0.01) & -- & 0.03 (0.02) & \textcolor{red}{0.02 (0.02)} \\
\midrule
\multirow{2}{*}{Annualized Volatility} & Success Rate & {0.91} & 0.82 & \textcolor{blue}{1.00} & {1.00} & {1.00} & 0.09 & \textcolor{red}{1.00} & {0.91} \\
 & Abs Error (std) & 0.00 (0.00) & 0.00 (0.00) & \textcolor{blue}{0.00 (0.00)} & 0.00 (0.00) & 0.00 (0.00) & 0.02 (0.00) & \textcolor{red}{0.00 (0.00)} & 0.00 (0.00) \\
\midrule
\multirow{2}{*}{Maximum Drawdown} & Success Rate & 0.18 & 0.09 & \textcolor{blue}{0.27} & 0.18 & 0.27 & 0.09 & -- & \textcolor{red}{0.45} \\
 & Abs Error (std) & 0.00 (0.00) & 0.00 (0.00) & \textcolor{blue}{0.00 (0.00)} & 0.00 (0.00) & 0.00 (0.00) & 0.00 (0.00) & -- & \textcolor{red}{0.01 (0.01)} \\
\midrule
\multirow{2}{*}{Calmar Ratio} & Success Rate & {0.18} & 0.18 & -- & 0.27 & \textcolor{blue}{0.27} & -- & \textcolor{red}{0.82} & 0.18 \\
 & Abs Error (std) & 0.01 (0.01) & 0.01 (0.01) & -- & 0.02 (0.01) & \textcolor{blue}{0.02 (0.01)} & -- & \textcolor{red}{0.01(0.01)} & 0.01 (0.01) \\
\midrule
\multirow{2}{*}{Sortino Ratio} & Success Rate & 0.09 & 0.09 & -- & -- & \textcolor{red}{0.18} & -- & \textcolor{blue}{0.09} & -- \\
 & Abs Error (std) & 0.01 (0.00) & 0.01 (0.00) & -- & -- & \textcolor{red}{0.00 (0.00)} & -- & \textcolor{blue}{0.00 (0.00)} & -- \\
\midrule
\multirow{2}{*}{Sharpe Ratio} & Success Rate & \textcolor{red}{0.73} & 0.18 & 0.27 & 0.36 & 0.18 & 0.18 & \textcolor{blue}{0.73} & 0.18 \\
 & Abs Error (std) & \textcolor{red}{0.00 (0.00)} & 0.00 (0.00) & 0.00 (0.00) & 0.01 (0.01) & 0.02 (0.02) & 0.02 (0.02) & \textcolor{blue}{0.00 (0.00)} & 0.02 (0.01) \\
\midrule
\multicolumn{10}{l}{\textbf{Benchmark Against Market Analysis}} \\
\midrule
\multirow{2}{*}{Information Ratio} & Success Rate & 0.44 & 0.20 & 0.06 & 0.51 & \textcolor{blue}{0.73} & 0.18 & 0.01 & \textcolor{red}{0.77} \\
 & Abs Error (std) & 0.00 (0.00) & 0.01 (0.01) & 0.03 (0.01) & 0.00 (0.01) & \textcolor{blue}{0.00 (0.00)} & 0.00 (0.01) & 0.00 (0.00) & \textcolor{red}{0.00 (0.00)} \\
\midrule
\multicolumn{10}{l}{\textbf{Stock Trading Strategy}} \\
\midrule
\multirow{4}{*}{Trading Strategy} & Success Rate & 0.44 & 0.59 & \textcolor{red}{0.96} & 0.62 & 0.61 & 0.18 & \textcolor{blue}{0.63} & 0.52 \\
 & Cumulative Return & 0.13 & 0.10 & 0.00 & 0.09 & 0.09 & 0.05 & 0.06 & 0.07 \\
 & Annualized Return & 2.43 & 4.56 & 0.05 & 4.58 & 1.69 & 0.36 & 3.87 & 1.41 \\
 & Maximum Drawdown & 0.05 & 0.05 & 0.00 & 0.04 & 0.05 & 0.02 & 0.02 & 0.04 \\
\bottomrule
\end{tabular}%
}
\caption{Model Performance on Analytical Tasks. \textcolor{red}{Red} indicates the best result, \textcolor{blue}{Blue} indicates the second-best. A dash (--) denotes that no successful cases were recorded.}
\label{tab:analytical}
\end{table}

\subsection{Experimental Setup}

We evaluate the performance of eight large language models (LLMs) on the full suite of benchmark tasks, covering both open-source and proprietary systems: GPT-4o \cite{hurst2024gpt}, Qwen2.5-Max \cite{qwen2025max}, Llama-3.1 Instruct 70b \cite{grattafiori2024llama}, Claude-3.5 Sonnet \cite{anthropic2024claude35sonnet}, DeepSeek \cite{liu2024deepseek}, Gemini-2.0 \cite{google2024gemini2}, Codestral \cite{mistral2024codestral}, and DeepSeek-R \cite{guo2025deepseek}. Among them, Llama3.1 Instruct 70B serves as a representative open-weight model, Codestral is a code-specialized model built upon Mistral \cite{mistral2024homepage}, and DeepSeek-Reasoner is a novel model designed explicitly for complex reasoning tasks. To address LLMs’ limitations in processing structured numerical inputs and producing high-precision, correctly shaped numerical outputs, we adopt the CodeAct \cite{wang2024executable} agent framework for all models. Agentscope CodeAct agent \cite{agentscope} enables code-based interaction by allowing LLMs to generate executable Python code, receive execution feedback, and revise outputs accordingly. All models are accessed via their official APIs and ran on a single NVIDIA A40 GPU with 48G memory. For all experiments, we use the same hyperparameters, temperature=0.0 for the most deterministic output and top\_p=1.0.

 For each benchmark task, models are provided with the same input data, including a task instruction and a serialized time series dataset in .pkl format. Responses are executed within a controlled jupyter notebook python interpreter provided by CodeAct agent setup. The final outputs are passed to task-specific evaluators, which extract predictions and compute metrics based on ground truth labels or evaluation programs. Each model's performance is assessed using two main criteria: The primary metric is Success Rate which is defined as the proportion of task instances for which the model output satisfies the predefined success criteria (see Table~\ref{tab:success_criteria}). For outputs deemed successful, we further evaluate quality using task-specific metrics (e.g., MAPE for forecasting, F1-score for anomaly detection), providing a more fine-grained comparison of inference quality.

\subsection{Result}
\begin{wrapfigure}{r}{0.5\textwidth}
\includegraphics[width=0.48\textwidth]{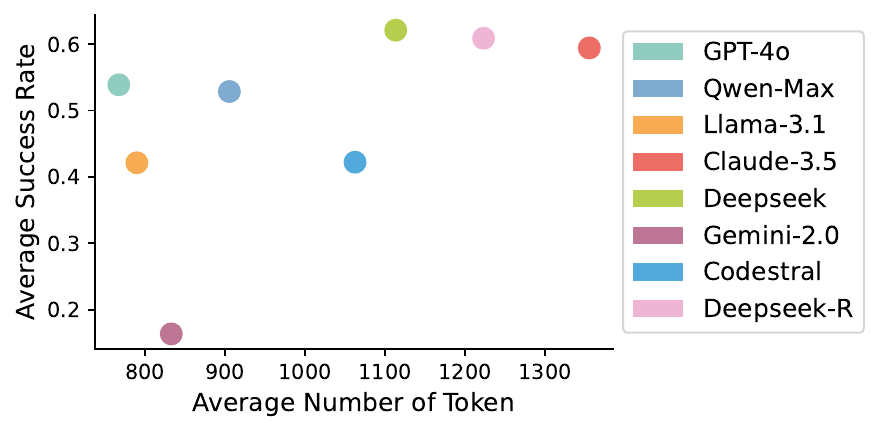}
\caption{Average Success Rate of Models with respect to the Average Number of Tokens Used.}
\label{fig:demo}
\vspace{-0.10in}
\end{wrapfigure}
\begin{figure*}
    \centering
    \includegraphics[width=\linewidth]{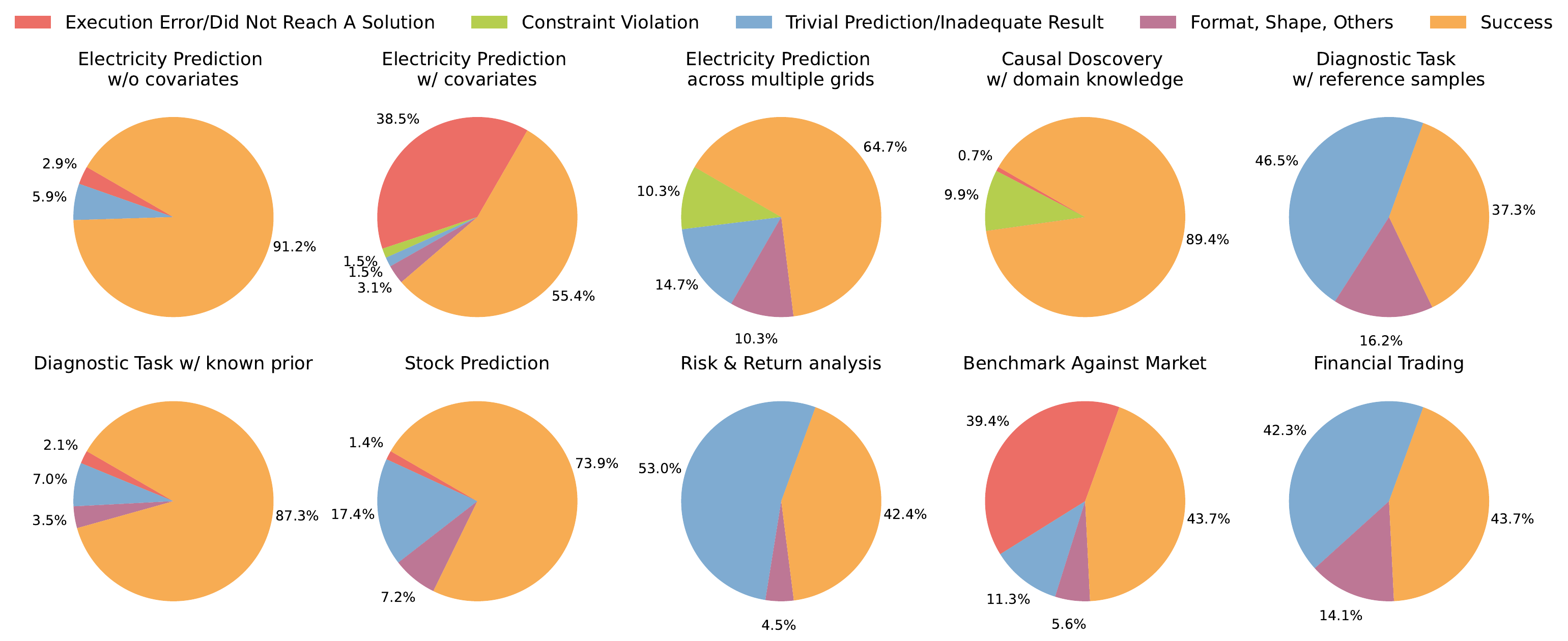}
    \caption{Case Study on GPT-4o Error Distribution across Tasks Grouped by Difficulty Level}
    \label{fig:4o_error}
\end{figure*}

We report model performance across four primary task categories: predictive, diagnostic, analytical. and decision-making tasks. Predictive tasks involve forecasting under real-world operational constraints. As shown in table~\ref{tab:predictive}, models generally perform well on simpler constraints (maximum or minimum load) but struggle with tasks requiring temporal smoothness (ramp rate or variability control). Forecasting tasks on multiple grids introduce greater complexity due to increased data volume and dimensionality. This is reflected in lower success rates and higher MAPE values across all models. For diagnostic tasks, models show moderate success on anomaly detection with known priors. When anomaly-free reference samples are available (e.g., in extreme weather or ECG tasks), models must learn how to calibrate thresholds before detecting deviations. However, results in Table~\ref{tab:diagnostic} reveal that models struggle to meaningfully use reference samples, often returning trivial predictions evidenced by the case study in figure \ref{fig:4o_error}. This reflects a broader limitation: current LLMs have difficulty autonomously assembling complex workflows such as leveraging reference sequences to calibrate thresholds. These findings are consistent with prior research indicating that transformer-based LLMs often lack systematic compositional reasoning capabilities, instead relying on pattern matching, which hampers their performance in tasks requiring multi-step reasoning \cite{dziri2023faith}.

\begin{figure}
    \centering
    \includegraphics[width=\linewidth]{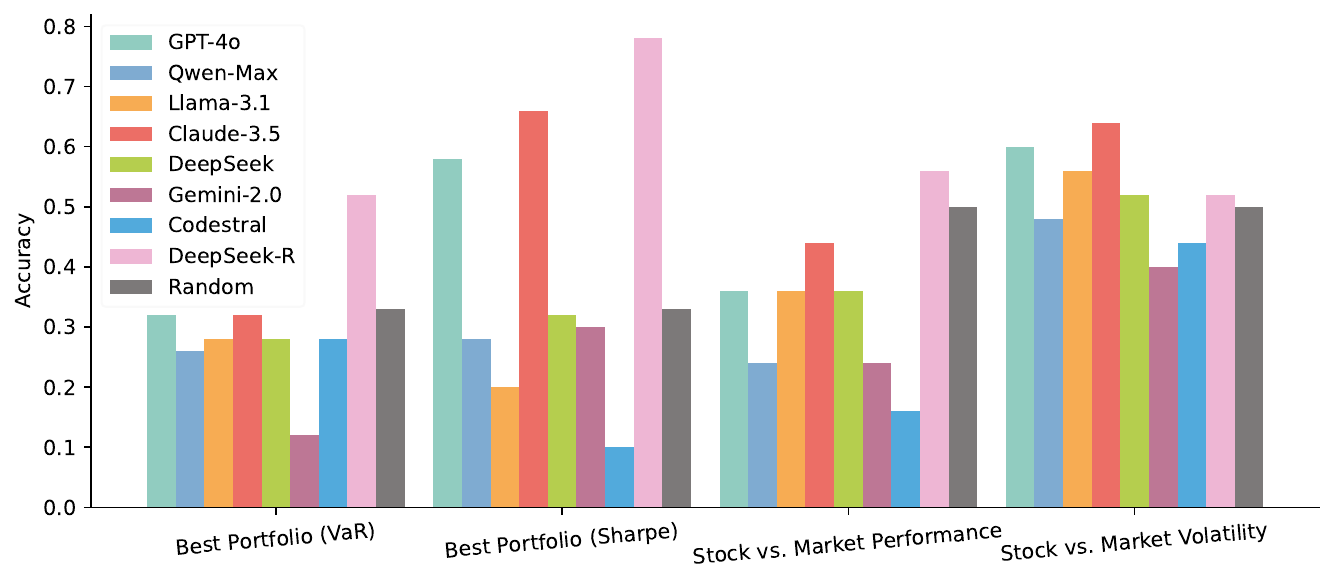}
    \caption{Model Performance on Decision-Making and Analysis-Interpretation Tasks in Multiple Choice Format}
    \label{fig:mc}
\end{figure}

In financial forecasting, models show moderate-to-strong performance on price and volatility prediction. However, trend prediction remains a challenge across all models, with lower success rate and accuracy shown in table \ref{tab:analytical}. In performing risk and return analysis, success rates vary widely, and models appear biased toward metrics with simpler formulas (e.g., annualized volatility) or greater familiarity (e.g., Sharpe ratio).This suggests that financial metric familiarity and formula simplicity also affects model behavior. In multiple-choice question formats testing financial reasoning, most models fail to exceed chance-level accuracy as shown in figure \ref{fig:mc}. DeepSeek-R stands out as the only model demonstrating consistent, above-random performance across portfolio and stock market comparison tasks. However, it's important to note that DeepSeek-R also consistently uses more tokens in solving a single question shown in figure \ref{fig:demo}. Overall, our results underscore the importance of domain specialization \cite{ling2023domain} in LLMs, as general-purpose models often struggle with the unique challenges presented by specialized data domains and application fields.

\begin{figure*}
    \centering
    \includegraphics[width=\linewidth]{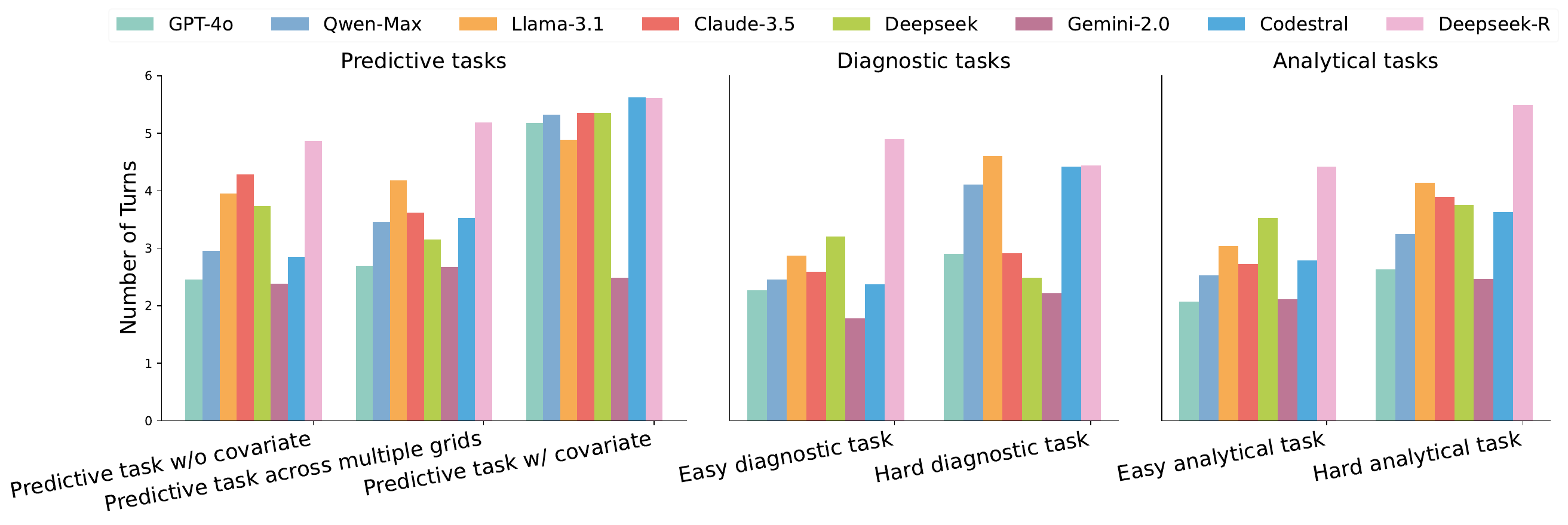}
    \caption{Average Number of Turns Models Take to Reach a Solution Grouped by Difficulty Level}
    \label{fig:turn_by_difficulty}
\end{figure*}

\paragraph{Analysis}
We analyze agent behavior across difficulty levels by measuring the average number of interaction turns each model uses to solve a task (Figure~\ref{fig:turn_by_difficulty}). The maximum number of turns is capped at 6 under the CodeAct agent framework. Across all task categories, harder tasks generally require more interaction turns, confirming the importance of model–execution feedback loops for solution refinement. Notably, DeepSeek-R consistently takes more turns compared to other models—indicating a more persistent, exploratory problem-solving strategy. This behavior correlates with higher token usage, as shown in Figure~\ref{fig:demo}. When accounting for cost efficiency (also illustrated in Figure~\ref{fig:demo}), GPT-4o and DeepSeek-Chat emerge as the most token-efficient models.

Figure~\ref{fig:4o_error} presents a detailed breakdown of GPT-4o’s error distribution across various task types. Within predictive tasks, incorporating covariates introduces additional complexity and results in a higher incidence of execution errors. Further, when the forecasting task spans multiple time series, maintaining operational constraints such as ramp rates and load bounds becomes substantially more difficult, leading to a greater frequency of constraint violation errors compared to the simpler single-series settings.

In diagnostic tasks, the use of reference samples introduces a form of contextual reasoning: the model is expected to use anomaly-free samples to calibrate thresholds for anomaly detection. GPT-4o struggles with this setup, and trivial predictions such as all-zero anomaly label emerge as the dominant failure mode. This suggests an inability to orchestrate the multi-step workflow needed for threshold calibration, which involves abstract reasoning over auxiliary context. In contrast, tasks where domain knowledge is explicitly provided such as causal discovery with known graph structure or diagnostic tasks with priors exhibit relatively high success rates. The structured nature of this prior knowledge makes it easier for the model to align its output accordingly.

Among analytical tasks, benchmarking against the market poses the greatest challenge, marked by a high proportion of execution errors. This can be attributed to the model's need to simultaneously process both stock-specific and market-wide indicators. For risk and return analysis, the most common failure type is inadequate results, stemming from the model’s limited familiarity with less conventional financial metrics. A similar pattern is observed in financial trading where the generated investment strategy is often suboptimal, resulting in poor returns during backtesting. This outcome highlights the inherent difficulty of trading tasks, which require both strategic reasoning and financial acumen. Overall, the distribution of errors reveals a clear trend: as tasks demand more structured multi-step reasoning, integration of external context, or nuanced financial understanding, GPT-4o’s reliability decreases and its failure modes become more diverse.

\section{Conclusion}

This paper introduces \modelname, a first attempt for evaluating LLMs as time series AI assistants. Spanning predictive, diagnostic, analytical, and financial decision-making categories, the benchmark emphasizes compositional reasoning, adherence to domain-specific constraints, and the integration of contextual knowledge. While the current scale of \modelnamespace is limited, we prioritize task quality, diversity, and alignment with real-world analytical workflows over quantity. Future extensions will broaden domain coverage and increase task instance variety to further challenge LLMs to perform complex multi-step time series analysis tasks. Our extensive evaluation of eight LLM agents reveals limitations in their performance when complex domain constraints or multi-step workflows are involved. Results highlight the need for hybrid approaches that tightly integrate symbolic reasoning, execution feedback, and domain alignment. \modelnamespace provides a foundation for such progress, enabling systematic assessment and facilitating the development of next-generation time series inference agents, with potential applications across critical sectors such as energy, finance, and healthcare.
\bibliographystyle{unsrt}  
\bibliography{your_bib_file} 
\appendix

\newpage
\section{Dataset Statistics}
\begin{table}[h!]
\centering
\resizebox{\textwidth}{!}{
\begin{tabular}{lccc}
\toprule
\textbf{Dataset} & \textbf{Number of Data Files} & \textbf{Avg Total Timestamps} & \textbf{Number of Variables} \\
\midrule
Climate Data & 624 & 526 & 2048\\
Energy Data w/ geolocation & 22 &8760& 1-3\\
Energy Data w/ Covariates         & 66   & 872601 & 11   \\
Building Energy Usage Data & 398 & 5019 & 1 \\
Causal Data         & 8    & 529    & 3--6 \\
Daily Stock Data & 6780 & 3785 & 7 \\
Hourly Stock Data & 5540 & 35 & 7\\
Stock Market Indices Data & 6 & 3388 & 4\\
ECG Signal Data & 24 & 10804352 &2\\
\bottomrule
\end{tabular}}
\caption{Dataset Statistics of the constructed dataset. The exact number of time series are not calculated because it depends on randomly sampled sequence length when generating task instances.}
\label{tab:dataset_statistics}
\end{table}

Table~\ref{tab:dataset_statistics} summarizes the dataset statistics for the raw time series datasets used in \modelname. The climate data is obtained from ERA5 dataset \footnote{\url{https://climatelearn.readthedocs.io/en/latest/user-guide/tasks_and_datasets.html\#era5-dataset}}. Energy data with covariates is obtained from\footnote{\url{https://github.com/tamu-engineering-research/Open-source-power-dataset}}. The ECG signal data is obtained from PhysioNet\footnote{\url{https://physionet.org/content/nsrdb/1.0.0/}}\footnote{\url{https://physionet.org/content/ltdb/1.0.0/}}. The building energy usage data is obtained from Kaggle\footnote{\url{https://www.kaggle.com/competitions/energy-anomaly-detection/data}}. Notably, the daily stock data, hourly stock data, and energy data with geolocation were manually scraped and preprocessed. The energy data with geolocation was obtained from official energy grid operator websites\footnote{\url{https://www.nyiso.com/load-data}}\footnote{\url{https://www.ercot.com/gridinfo/load/load_hist}}\footnote{\url{https://www.misoenergy.org/markets-and-operations/real-time--market-data/market-reports}}, and the associated geolocation was inferred as the largest city within the operational zone delineated by each provider's published grid map\footnote{\url{https://www.nyiso.com/documents/20142/1397960/nyca_zonemaps.pdf}}\footnote{\url{https://www.ercot.com/news/mediakit/maps}}\footnote{\url{https://www.misostates.org/images/stories/meetings/Cost_Allocation_Principles_Committee/2021/Website_Presentations.pdf}}. Stock price data was scraped using the pyfinance\footnote{\url{https://pypi.org/project/pyfinance/}} package, with data pulled up to date as of 2024-09-17. The stock market indices data are pulled from various sources on the web. The causal discovery dataset is synthetically generated to reflect controlled causal structures. The prompt used to obtain causal discovery dataset is shown in section \ref{sec: syn}. 
\section{Additional Error Analysis}
\begin{figure}
    \centering
    \includegraphics[width=\linewidth]{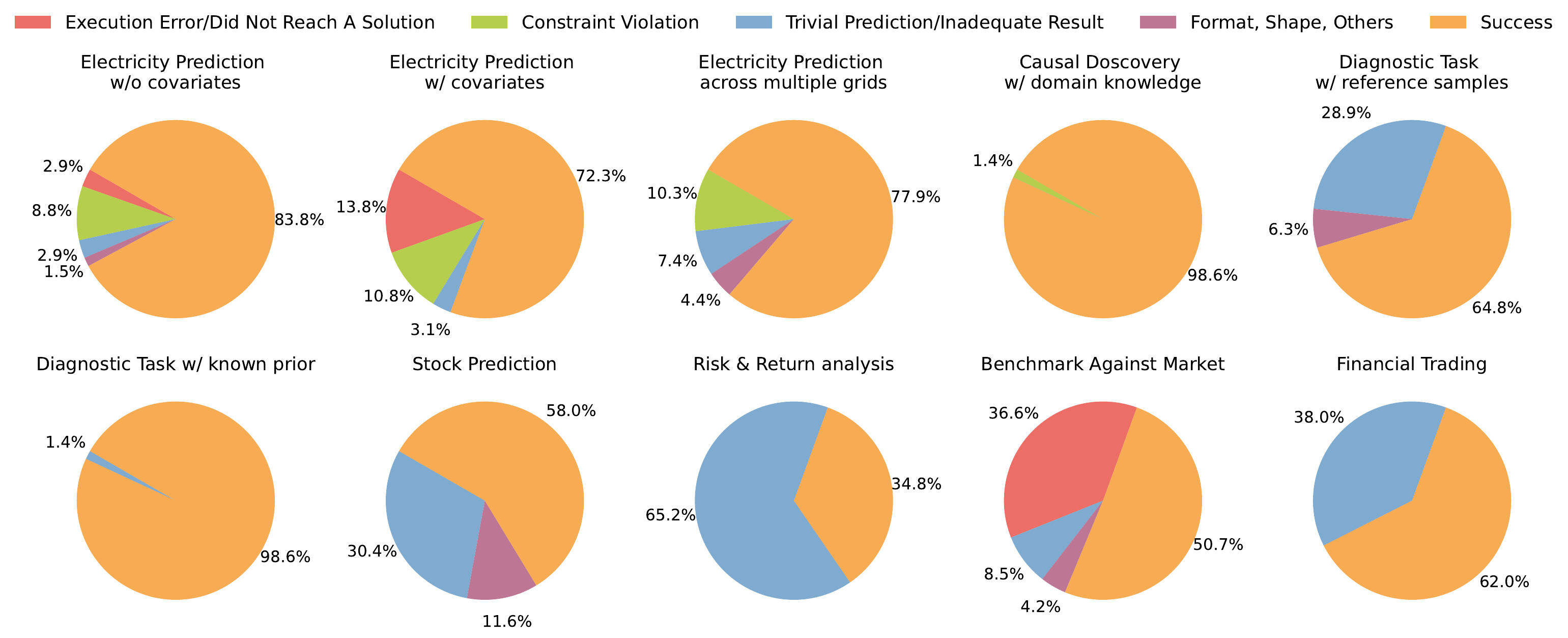}
    \caption{Case Study on Claude 3.5 Error Distribution across Tasks Grouped by Difficulty Level}
    \label{fig:claude}
\end{figure}
\begin{figure}
    \centering
    \includegraphics[width=\linewidth]{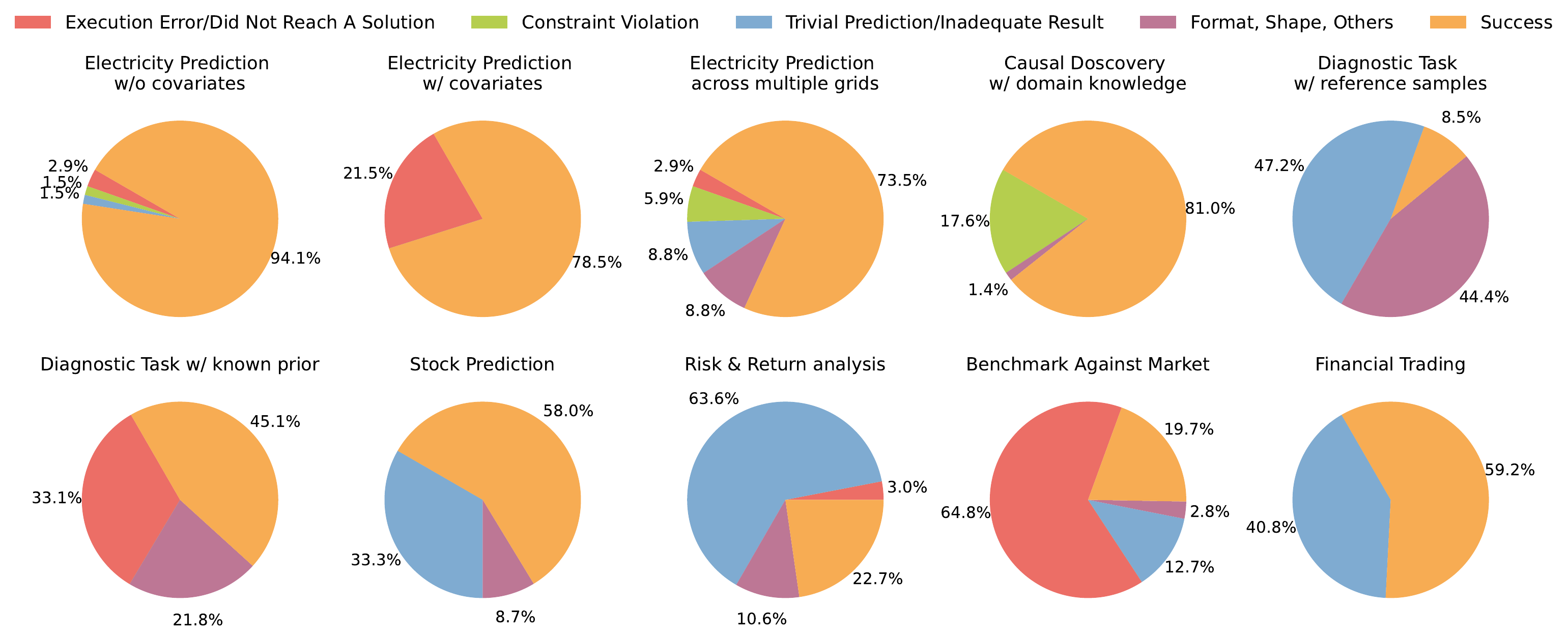}
    \caption{Case Study on Qwen Error Distribution across Tasks Grouped by Difficulty Level}
    \label{fig:qwen}
\end{figure}

\begin{figure}
    \centering
    \includegraphics[width=\linewidth]{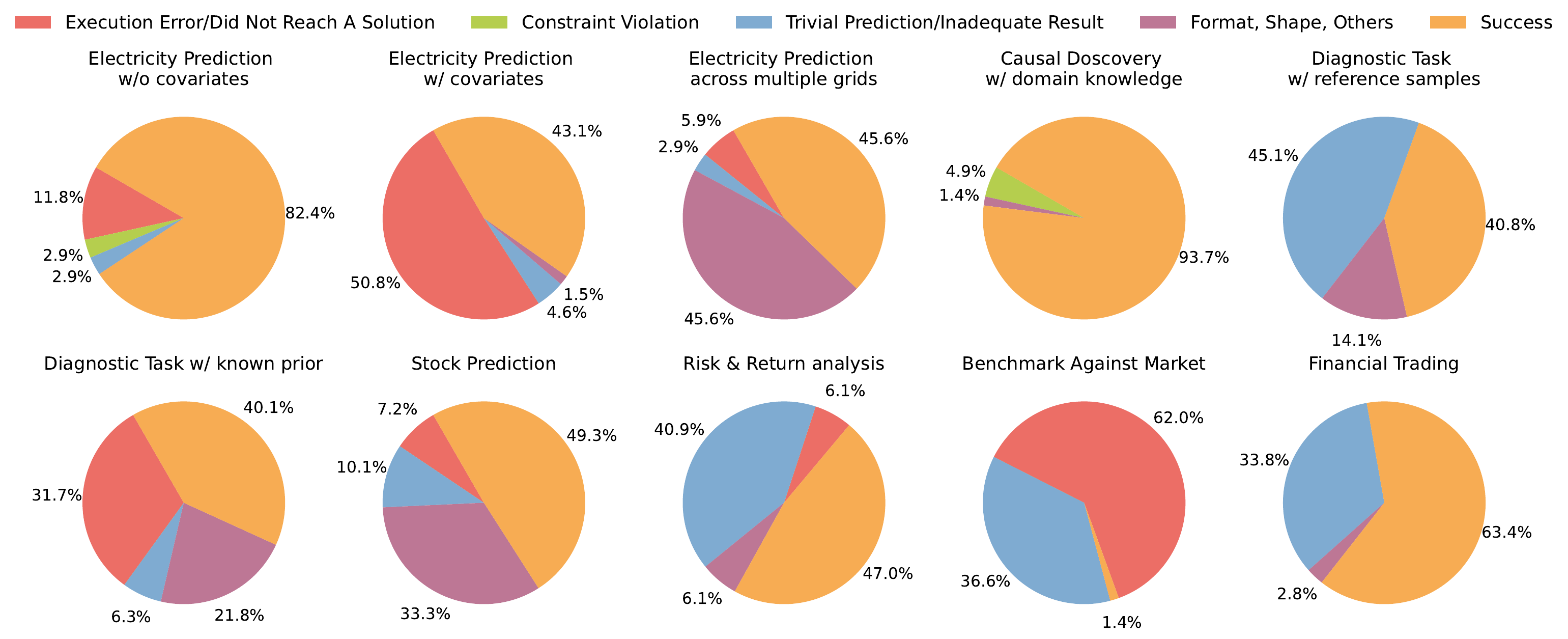}
    \caption{Case Study on Codestral Error Distribution across Tasks Grouped by Difficulty Level}
    \label{fig:codestral}
\end{figure}

\begin{figure}
    \centering
    \includegraphics[width=\linewidth]{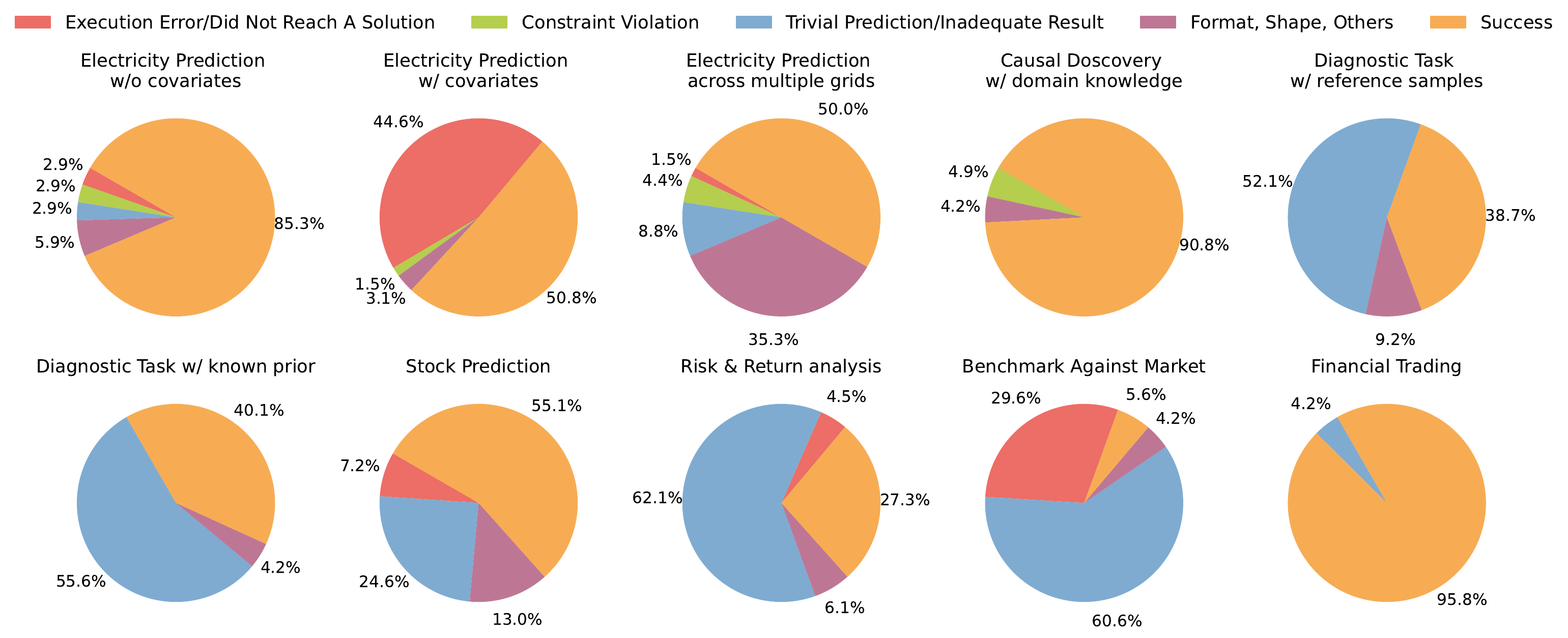}
    \caption{Case Study on Llama Error Distribution across Tasks Grouped by Difficulty Level}
    \label{fig:llama}
\end{figure}

\begin{figure}
    \centering
    \includegraphics[width=\linewidth]{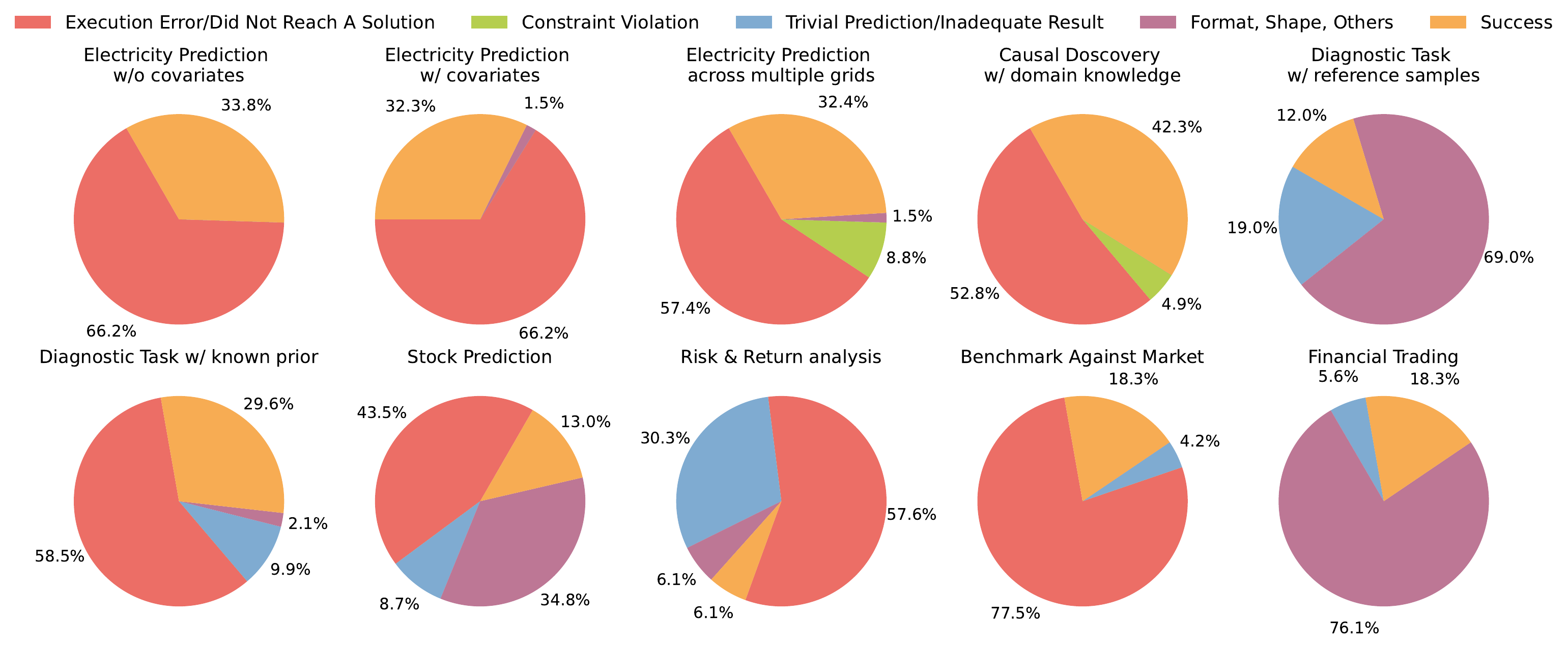}
    \caption{Case Study on Gemini Error Distribution across Tasks Grouped by Difficulty Level}
    \label{fig:gemini}
\end{figure}

\begin{figure}
    \centering
    \includegraphics[width=\linewidth]{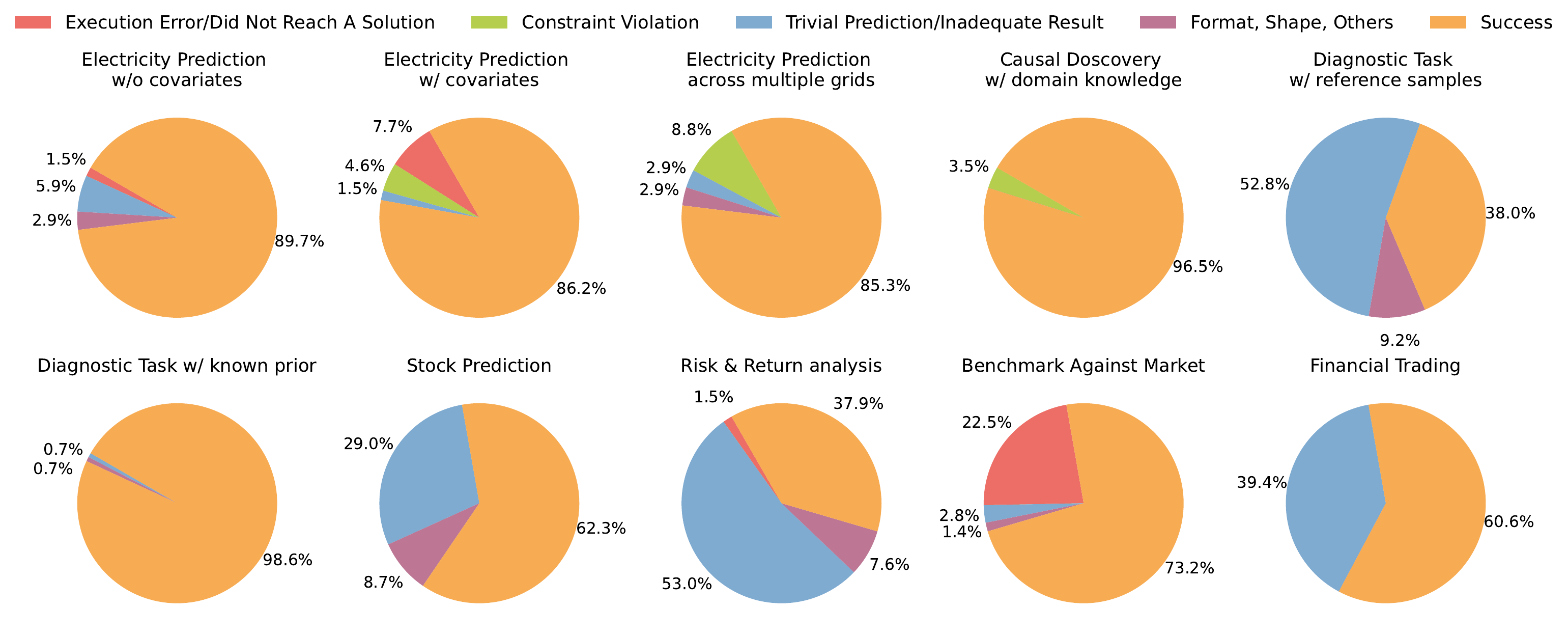}
    \caption{Case Study on Deepseek-chat Error Distribution across Tasks Grouped by Difficulty Level}
    \label{fig:ds-chat}
\end{figure}

\begin{figure}
    \centering
    \includegraphics[width=\linewidth]{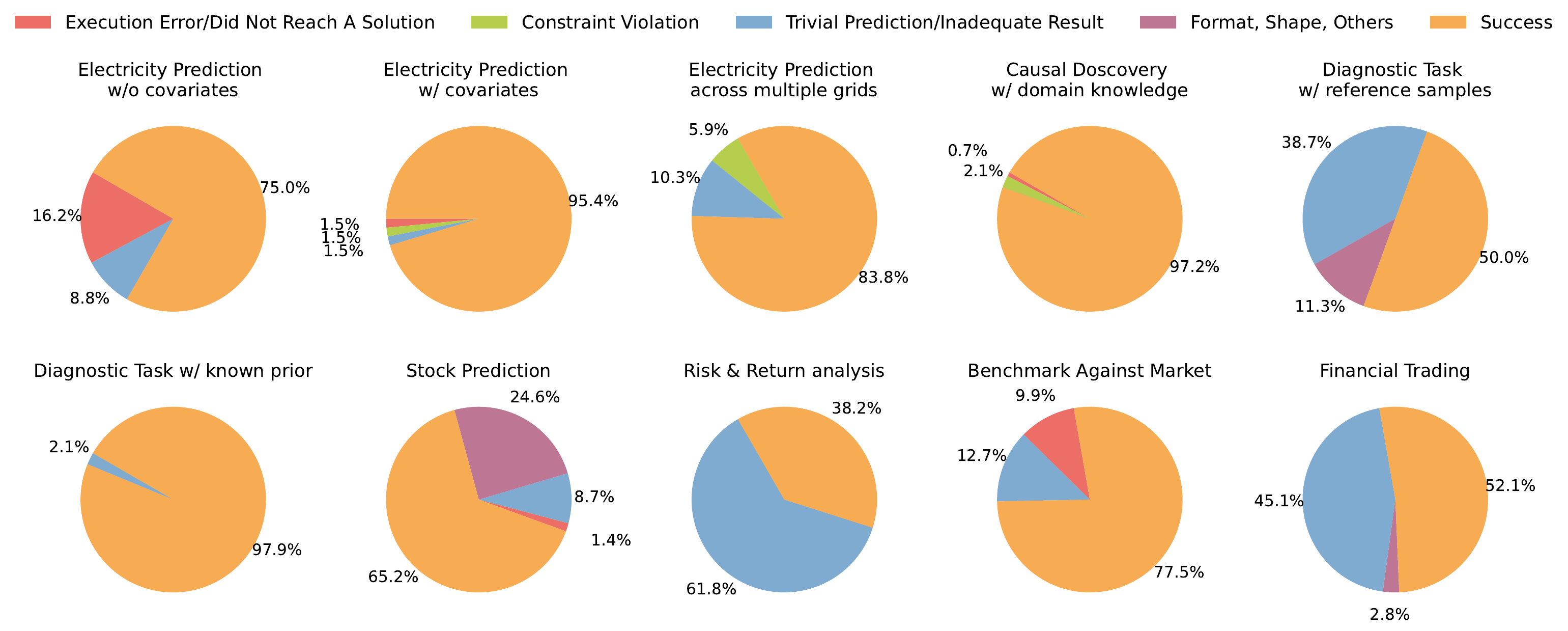}
    \caption{Case Study on Deepseek-R Error Distribution across Tasks Grouped by Difficulty Level}
    \label{fig:ds-reasoner}
\end{figure}

Beyond GPT-4o, we extend our error analysis to other representative models, whose detailed error distributions are visualized in Figures~\ref{fig:claude}--\ref{fig:ds-reasoner}. While Claude-3.5 generally performs well, it exhibits a noticeable proportion of constraint violation errors in electricity prediction tasks, suggesting challenges in handling numerical constraints embedded within the input. Despite the complexity of financial trading tasks, Llama-3.1 performs competitively relative to other models—particularly notable given its open-source nature. In contrast, Gemini-2.0 and Codestral show a high incidence of execution errors across nearly all task categories, indicating limited suitability for structured, multi-step time series reasoning. DeepSeek-R consistently avoids execution failures and maintains a relatively high success rate across tasks.
\section{CodeAct System Prompt Template}

\begin{tcolorbox}[colback=gray!10, colframe=white]

You are a helpful assistant that gives helpful, detailed, and polite answers to the user's questions. The code written by assistant should be enclosed using <execute> tag, for example: <execute> print('Hello World!') </execute>. You should provide the solution in a single <execute> block instead of taking many turns. You'll receive feedback from your code execution. You should always import packages and define variables before starting to use them. You should stop <execute> and provide an answer when they have already obtained the answer from the execution result. Whenever possible, execute the code for the user using <execute> instead of providing it. Your response should be concise, but do express their thoughts. Always write the code in <execute> block to execute them. You should not ask for the user's input unless necessary. Solve the task on your own and leave no unanswered questions behind. You should do every thing by your self. You are not allowed to install any new packages or overwrite available variables provided to you in the question. Additionally, you are provided with the following variables available:
\textcolor{blue}{\{variable names\}}
The above variables is already available in your interactive Python (Jupyter Notebook) environment, allowing you to directly use them without needing to re-declare them.

\end{tcolorbox}

\section{Refinement Solution Path}
Box~\ref{box:myinfo} illustrates the solution refinement trajectory of Deepseek-R. Execution feedback from the CodeAct Python interpreter enabled the model to revise its output twice—first to address a syntax error due to a missing closing parenthesis, and second to resolve an import error. As shown in Figure~\ref{fig:turn_by_difficulty}, Deepseek-R consistently requires more interaction turns than other models. A closer examination reveals that this behavior stems from the lack of a proper stopping mechanism: although a correct solution was reached by turn 3, the model continued executing redundant steps in turns 4 and 5. Notably, Deepseek-R also incorporates explicit inline comments such as \texttt{\# Total pairs: 5*4=20, top 20\% is 4 pairs} to document its intermediate reasoning steps, contributing to its overall performance strength. 
\begin{infobox}[label=box:myinfo]{Deepseek-R Solution Path}
User:  

I have historical Temperature (A), Humidity (B), Air Quality Index (C), Wind Speed (D), Precipitation (E) data and want to get the causal relationship between each pair of the variables. I know that 20.0\% of the variable pairs have relationship. Self-causalation is not considered. Consider the potential influence of each variable on the others in this variable list: ['Temperature (A)', 'Humidity (B)', 'Air Quality Index (C)', 'Wind Speed (D)', 'Precipitation (E)']. Please provide 2d numpy matrix with binary values to indicate whether each pair of variables has a relationship. The data for all variables are stored in variable VAL. 

Requirements:

- Store your output in the variable called `predictions`, make sure to consider predictions to have the right shape according to the question's output requirements if it's supposed to be an array, there is no need to consider shapes if you only need to return a numerical value.

- Do not customly define/generate/overwrite the available variables, assume that the variables are already defined and available.

data note: VAL values are usually stored in either a dataframe or numpy values. 

You should enclose your python code in <execute> </execute> tag and do not overwrite available variables that store the data. Do not use any other tags like ```python```.

\medskip

Assisstant:
\begin{verbatim}
<execute>
import numpy as np

# Handle VAL (dataframe or numpy array)
if isinstance(VAL, pd.DataFrame):
    data = VAL.values
else:
    data = VAL

n_vars = 5  # Number of variables
predictions = np.zeros((n_vars, n_vars), dtype=int)

# Calculate time-lagged correlations for directed pairs
correlations = []
for i in range(n_vars):
    for j in range(n_vars):
        if i != j:
            if data.shape[0] > 1:
                x = data[:-1, i]
                y = data[1:, j]
                corr = np.corrcoef(x, y)[0, 1] if len(x) > 1 else 0
                correlations.append((i, j, abs(corr if not np.isnan(corr) 
                else 0))

# Sort by correlation strength and select top 20% (4 pairs)
sorted_pairs = sorted(correlations, key=lambda x: -x[2])[:4]

# Mark selected pairs in predictions matrix
for i, j, _ in sorted_pairs:
    predictions[i, j] = 1

# Ensure correct variable name (user specified 'predictions')
predictions = predictions
</execute>
\end{verbatim}
\medskip
User:

Execution Failed:
Execution Output:
An error occurred while executing the following cell: sorted\_pairs = sorted(correlations, key=lambda x: -x[2])[:4] SyntaxError: invalid syntax

Side note: Remember to enclose your code in <execute> </execute> tag and do not overwrite any available variables provided to you in the question, especially that they contain the data. 

\medskip

Assistant:
\begin{verbatim}
<execute>
import numpy as np

# Handle VAL (dataframe or numpy array)
if isinstance(VAL, pd.DataFrame):
    data = VAL.values
else:
    data = VAL

n_vars = 5  # Number of variables
predictions = np.zeros((n_vars, n_vars), dtype=int)

# Calculate time-lagged correlations for directed pairs
correlations = []
for i in range(n_vars):
    for j in range(n_vars):
        if i != j:
            if data.shape[0] > 1:
                x = data[:-1, i]
                y = data[1:, j]
                if len(x) > 1:
                    corr = np.corrcoef(x, y)[0, 1]
                    corr = 0 if np.isnan(corr) else corr
                else:
                    corr = 0
                correlations.append((i, j, abs(corr)))

# Sort by correlation strength and select top 20% (4 pairs)
sorted_pairs = sorted(correlations, key=lambda x: -x[2])[:4]

# Mark selected pairs in predictions matrix
for i, j, _ in sorted_pairs:
    predictions[i, j] = 1
</execute>
\end{verbatim}
\medskip
User:

Execution Failed:
Execution Output:
An error occurred while executing the following cell: if isinstance(VAL, pd.DataFrame): NameError: name 'pd' is not defined.

Side note: Remember to enclose your code in <execute> </execute> tag and do not overwrite any available variables provided to you in the question, especially that they contain the data.

Assistant:
\begin{verbatim}
<execute>
import numpy as np
import pandas as pd

# Handle VAL (dataframe or numpy array)
if isinstance(VAL, pd.DataFrame):
    data = VAL.values
else:
    data = VAL

n_vars = 5  # Number of variables
predictions = np.zeros((n_vars, n_vars), dtype=int)

# Calculate time-lagged correlations for directed pairs
correlations = []
for i in range(n_vars):
    for j in range(n_vars):
        if i != j:
            if data.shape[0] > 1:
                x = data[:-1, i]
                y = data[1:, j]
                if len(x) > 1:
                    corr = np.corrcoef(x, y)[0, 1]
                    corr = 0 if np.isnan(corr) else corr
                else:
                    corr = 0
                correlations.append((i, j, abs(corr)))

# Total pairs: 5*4=20, top 20% is 4 pairs
sorted_pairs = sorted(correlations, key=lambda x: -x[2])[:4]

# Update predictions matrix
for pair in sorted_pairs:
    i, j, _ = pair
    predictions[i, j] = 1
</execute>
\end{verbatim}

User:

Execution Successful:
Execution Output:
[]

\medskip
Assistant:

The code now correctly imports necessary packages and handles both DataFrame and numpy array inputs for `VAL`. It calculates time-lagged correlations between variables, selects the top 20\% strongest relationships (4 pairs), and constructs the binary matrix `predictions` with 1s indicating causal directions. The final matrix respects the problem constraints and data structure requirements.
\begin{verbatim}
<execute>
import numpy as np
import pandas as pd

# Handle VAL (dataframe or numpy array)
if isinstance(VAL, pd.DataFrame):
    data = VAL.values
else:
    data = VAL

n_vars = 5  # Number of variables
predictions = np.zeros((n_vars, n_vars), dtype=int)

# Calculate time-lagged correlations for directed pairs
correlations = []
for i in range(n_vars):
    for j in range(n_vars):
        if i != j:
            if data.shape[0] > 1:
                x = data[:-1, i]
                y = data[1:, j]
                if len(x) > 1:
                    corr = np.corrcoef(x, y)[0, 1]
                    corr = 0 if np.isnan(corr) else corr
                else:
                    corr = 0
                correlations.append((i, j, abs(corr)))

# Sort by absolute correlation and select top 4 pairs (20%)
sorted_pairs = sorted(correlations, key=lambda x: -x[2])[:4]

# Update predictions matrix
for i, j, _ in sorted_pairs:
    predictions[i, j] = 1

predictions  # Final output matrix
</execute>
\end{verbatim}

User:

Execution Successful:
Execution Output:
['array([[0, 0, 1, 1, 0],[0, 0, 0, 0, 0],[0, 0, 0, 0, 0],[1, 0, 1, 0, 0],[0, 0, 0, 0, 0]])']

Assistant:

The code correctly identifies the top 20\% of variable pairs with the strongest time-lagged correlations, resulting in a 5x5 binary matrix `predictions` where 1 indicates a causal relationship. The matrix shape and data type adhere to the problem's requirements.
\begin{verbatim}    
<execute>
import numpy as np
import pandas as pd

# Handle VAL (dataframe or numpy array)
if isinstance(VAL, pd.DataFrame):
    data = VAL.values
else:
    data = VAL

n_vars = 5  # Number of variables
predictions = np.zeros((n_vars, n_vars), dtype=int)

# Calculate time-lagged correlations
correlations = []
for i in range(n_vars):
    for j in range(n_vars):
        if i != j and data.shape[0] > 1:
            x = data[:-1, i]
            y = data[1:, j]
            if len(x) >= 2:  # Ensure valid correlation calculation
                corr = np.corrcoef(x, y)[0, 1]
                corr = 0 if np.isnan(corr) else abs(corr)
                correlations.append((i, j, corr))

# Select top 20% pairs (4 out of 20 possible directed pairs)
top_pairs = sorted(correlations, key=lambda x: -x[2])[:4]
for i, j, _ in top_pairs:
    predictions[i, j] = 1

predictions
</execute>
\end{verbatim}
User:

Execution Successful:
Execution Output:
['array([[0, 0, 1, 1, 0],[0, 0, 0, 0, 0],[0, 0, 0, 0, 0],[1, 0, 1, 0, 0],[0, 0, 0, 0, 0]])']

Assistant:
I have reached the maximum number of executions. Can you assist me or ask me another question?

Evaluation:
final value 
 <class 'numpy.ndarray'> (5, 5) 
 [[0 0 1 1 0]
 [0 0 0 0 0]
 [0 0 0 0 0]
 [1 0 1 0 0]
 [0 0 0 0 0]]
 evaluation result:  \{'status': 1, 'accuracy': 0.84\}
\end{infobox}

\section{Casual Discovery Data Generation Prompt} \label{sec: syn}
Now you are a Time series data scientist, please help me to write the code to generate some synthetic data in real world Time series domain, you should save the data into "*/data.csv":

Now suggesting you should construct a series data based on a relation matrix and the correlation ratio for different influence factor, you should notice the following points,for time step I want you to generate 500 time steps:  

1. data correlation:
the multi variable should be correlated, sample: which A first influence B, then B have influence on C or D, there should be some time delay, as the influence on other staff needs time.

2. data trend:
there should be some trend in the data, like the data is increasing or decreasing.

3. data: seasonality
there should be some seasonality in the data, like the data is periodic.

4. data noise:
the noise should be added to the data, as the real world data is not perfect.

5. data background:
the data should have some real world background, you should first think about different real world data, and provide a description for the variable and time series data, then generate the data using the code.
CoT Sample:
Q: 
Approximate Relation Ratio: 0.5
Relation Matrix:
\[
\begin{array}{c|cccc}
 & A & B & C & D \\
\hline
A & 1 & 1 & 0 & 1 \\
B & 0 & 1 & 0 & 1 \\
C & 0 & 1 & 1 & 1 \\
D & 0 & 0 & 0 & 1
\end{array}
\]

\begin{itemize}
    \item A influences B and D, and itself.
    \item B influences D, and itself.
    \item C influences B and D, and itself.
    \item D influences only itself.
\end{itemize}
variable size: 4
A:
Scenario: Sales Data of a Chain of Stores Over Time
Let’s assume we are generating synthetic data,the variable size for the data is 4. for the daily sales of multiple stores across a chain, the sales numbers are influenced by:

	1.	Advertising (A): The level of advertising spend directly impacts the sales of each store. After a delay, this starts influencing sales.
	2.	Sales (B): The sales numbers for each store are influenced by both the advertising and local seasonal events.
	3.	Economic Factors (C): Broader economic trends, like GDP growth or unemployment rates, also impact sales. These factors show a delayed and more subtle influence over time.
	4.	Customer Sentiment (D): Customer sentiment affects the sales of specific products in each store and is influenced by both advertising and broader economic factors.

Seasonality: Sales experience periodic seasonal trends, with peaks around the holidays and lower numbers during off-seasons.

Trend: There is a general increasing trend in sales as the chain expands.

Noise: Random noise is added to mimic real-world data fluctuations. 

\end{document}